%% file: ms.tex
\documentclass{article}

    \PassOptionsToPackage{numbers}{natbib}

    \usepackage[final]{neurips_2020}

\usepackage[utf8]{inputenc} 
\usepackage[T1]{fontenc}    
\usepackage{hyperref}       
\usepackage{url}            
\usepackage{booktabs}       
\usepackage{amsfonts}       
\usepackage{nicefrac}       
\usepackage{microtype}      

\usepackage{subfiles}
\usepackage{amsmath}
\usepackage{subfig}
\usepackage[pdftex]{graphicx}
\usepackage{algorithm,algorithmic}

\newcommand{\argmax}{\mathop{\rm arg~max}\limits}

\title{Utilizing Skipped Frames in Action Repeats\\ via Pseudo-Actions}

\author{%
  Taisei~Hashimoto \hspace{10mm} Yoshimasa~Tsuruoka \\
  Department of Information and Communication Engineering\\
  Graduate School of Information Science and Technology\\
  The University of Tokyo\\
  \texttt{\{hashi,tsuruoka\}@logos.t.u-tokyo.ac.jp}
}

\begin{document}

\maketitle

\subfile{abstract}

\subfile{introduction}

\subfile{background}

\subfile{proposal}

\subfile{experiment}

\subfile{conclusion}

\subfile{broader_impact}

\bibliographystyle{plainnat}
\bibliography{ms}

\newpage

\subfile{appendix}

\end{document}

%% file: abstract.tex
\begin{abstract}
  In many deep reinforcement learning settings, when an agent takes an action, it repeats the same action a predefined number of times without observing the states until the next action-decision point.
  This technique of action repetition has several merits in training the agent, but the data between action-decision points (i.e., intermediate frames) are, in effect, discarded.
  Since the amount of training data is inversely proportional to the interval of action repeats, they can have a negative impact on the sample efficiency of training.
  In this paper, we propose a simple but effective approach to alleviate to this problem by introducing the concept of pseudo-actions.
  The key idea of our method is making the transition between action-decision points usable as training data by considering pseudo-actions.
  Pseudo-actions for continuous control tasks are obtained as the average of the action sequence straddling an action-decision point.
  For discrete control tasks, pseudo-actions are computed from learned action embeddings.
  This method can be combined with any model-free reinforcement learning algorithm that involves the learning of Q-functions.
  We demonstrate the effectiveness of our approach on both continuous and discrete control tasks in OpenAI Gym.
\end{abstract}

%% file: introduction.tex
\section{Introduction}
  \label{sec:introduction}

  In order to make reinforcement learning (RL) work in the real world, it is essential to have stable methods for learning from small amounts of data.
  Many studies have tackled this problem by improving algorithms, focusing on how to make better use of the data collected \citep{hesselRainbowCombiningImprovements2018,schulmanProximalPolicyOptimization2017,haarnojaSoftActorCriticOffPolicy2018}.
  By contrast, Kostrikov et al. \citep{kostrikovImageAugmentationAll2020} have shown that data augmentation, a commonly used technique in the field of computer vision, allows for efficient learning of a policy from images.
  Their results show that deep RL, like other fields, is susceptible to the negative effects of overfitting and that it can be mitigated by increasing the amount of data.
  Motivated by this observation, we explore another approach to increase the amount of training data for the agent, focusing on \textit{action repeats}, a technique commonly used in deep RL.

  In many RL settings, once an agent chooses an action, the action is repeated a fixed number of times.
  The use of this action repetition technique is known to have a significant impact on the quality of the learned policy \citep{hafnerDreamControlLearning2020,braylanFrameSkipPowerful2015}.
  In Atari 2600 \citep{machadoRevisitingArcadeLearning2017}, a common benchmark for discrete control, the number of repeated actions is often set to 4 \citep{machadoRevisitingArcadeLearning2017,kaiserModelBasedReinforcementLearning2020}.
  In the DeepMind control suite \citep{tassaDeepMindControlSuite2018}, a common benchmark for continuous control, repeats of 2 to 4 timesteps are used depending on the environments \citep{hafnerLearningLatentDynamics2019,leeStochasticLatentActorCritic2019}.

  Action repeats are widely adopted to stabilize training \citep{bellemareIncreasingActionGap2016} and to  facilitate exploration \citep{hesselInductiveBiasesDeep2019}.
  However, one problem with action repeats is that the intermediate frames between action-decision points are not fully utilized.
  In the DeepMind control suite, the data during action repeats are simply discarded.
  In Atari 2600, the last two frames during an action repeat are max-pooled to avoid overlooking objects that only appear at even or odd time steps.
  As long as the number of environment steps is constant, the amount of training data is inversely proportional to the interval of action repeats, which means a significant amount of data are discarded when the action repeat parameter is large.
  In this paper, we aim to utilize the information obtained in state transitions during action repeats effectively.

  The key idea of our method is to make use of the data during action repeats by pretending that a pseudo-action, which is computed as the average of actions, was repeated a designated number of times.
  While we can directly apply this method to the tasks with continuous action spaces, it is not readily applicable to the tasks with discrete action spaces.
  We address this issue by learning the embeddings of actions and to use their averages.

  The contributions of this paper are as follows:
  (1) We propose a method to increase the amount of data in continuous control tasks by making use of the transitions during action repeats.
  (2) We extend our method to discrete control tasks by learning the embeddings of actions.
  (3) By combining our method with Deep Q-learning \citep{mnihPlayingAtariDeep2013} and Soft Actor-Critic \citep{haarnojaSoftActorCriticOffPolicy2018}, we improve the performance in several continuous and discrete control tasks.


%% file: background.tex
\section{Background}
  \label{sec:background}

  \subsection{MDP}
    In RL, an environment is represented as a Markov decision process (MDP).
    An MDP is defined by $(\mathcal{S}, \mathcal{A}, p, r, \gamma)$, where $\mathcal{S}$ is the state space, $\mathcal{A}$ is the action space, $p$: $\mathcal{S}\times\mathcal{S}\times\mathcal{A}\rightarrow [0, \infty)$ is the transition dynamics, $r$: $\mathcal{S}\times\mathcal{A}\rightarrow\mathbb{R}$ is the reward function, and $\gamma\in[0, 1)$ is a discount factor.

    In this work, we conduct experiments using images as observations.
    In this case, the environment is formulated as a partially observable Markov decision process (POMDP) since some information about the state (e.g. the velocity of an object) cannot be obtained from a single image.
    A POMDP is described as $(\mathcal{O}, \mathcal{A}, p, r, \gamma)$, where $\mathcal{O}$ is the observation space.
    Following the convention \citep{mnihPlayingAtariDeep2013}, we convert a POMDP into an MDP by stacking several consecutive image observations into a state $s_t = \{o_t, o_{t-1}, o_{t-2}, \ldots\}$.

  \subsection{Deep Q-learning}
    Deep Q-learning \citep{mnihPlayingAtariDeep2013} is a well-known algorithm for discrete control tasks.
    The expected value of the cumulative discounted return obtained after taking an action $a$ in a state $s$ is called Q-value, and the mapping $Q(s,a)$ is called Q-function.
    In Deep Q-learning, the Q-function is estimated by a neural network $Q_\theta(s,a)$, where $\theta$ is a set of parameters.
    $Q_\theta(s,a)$ is trained by iteratively minimizing the following loss function.
    \begin{align*}
      L_Q(\mathcal{D}) &= \mathbb{E}_{s,a,s^\prime \sim \mathcal{D}} \left[ (y_i - Q_{\theta_i}(s,a))^2 \right], \\
      y_i &= r + \gamma \max_{a^\prime} Q_{\theta_{i-1}}(s^\prime, a^\prime),
    \end{align*}
    where $\mathcal{D}$ is a replay buffer for transition data, and $s^\prime$ is the next state of $s$.
    Due to action repetition, $s^\prime$ is the state that is reached by repeating action $a$ a certain number of times from state $s$.
    $\theta^\prime$ is a fixed old set of parameters, which is synchronized at regular intervals.

    Once Q-network $Q_\theta(s,a)$ is trained, the policy is obtained as follows:
    \begin{align*}
      a = \argmax_a Q_\theta(s,a) .
    \end{align*}

  \subsection{Soft Actor-Critic}
    Soft Actor-Critic (SAC) \citep{haarnojaSoftActorCriticOffPolicy2018,haarnojaSoftActorCriticAlgorithms2019} is a popular algorithm for continuous control tasks.
    In addition to a Q-network $Q_\theta(s,a)$, SAC trains a policy network $\pi_\phi(a|s)$.
    The policy is represented by a diagonal Gaussian distribution $\mathcal{N}(\mu_\phi(s), \sigma_\phi(s))$ followed by a $\tanh$ normalization.
    An action $a$ is sampled using a diagonal Gaussian noise $\varepsilon \sim \mathcal{N}(0, I)$: $a = \tanh(\mu_\phi(s) + \sigma_\phi(s) \odot \varepsilon)$.

    The Q-network $Q_\theta(s,a)$ is trained by optimizing the following loss function:
    \begin{align*}
      L_Q(\mathcal{D}) &= \mathbb{E}_{s,a,s^\prime \sim \mathcal{D}} \left[ (y_i - Q_{\theta}(s,a))^2 \right], \\
      y_i &= r + \gamma \mathbb{E}_{a^\prime \sim \pi(\cdot|s^\prime)} \left[ Q_{\theta^\prime}(s^\prime, a^\prime) - \alpha \log\pi_\theta(a^\prime|s^\prime) \right],
    \end{align*}
    where $\alpha$ is a temperature parameter, and $\theta^\prime$ is an exponential moving average of $\theta$.

    The policy $\pi(s,a)$ is trained by optimizing the objective
    \begin{align*}
      L_\pi(\mathcal{D}) &= \mathbb{E}_{s \sim \mathcal{D}} \left[ D_{\textrm{KL}} \left(\pi_\theta(\cdot|s_t) || \exp \left(\frac{1}{\alpha} Q(s,\cdot) \right)\right) \right].
    \end{align*}

    Finally, temperature $\alpha$ is adjusted so that the entropy of the policy gets close to the target value $\bar{H}$
    \begin{align*}
      L_\alpha (\mathcal{D}) = \mathbb{E}_{\substack{s\sim\mathcal{D}\\a\sim\pi_\theta(\cdot|s)}} \left[ -\alpha \log \pi_\theta(a|s) - \alpha \bar{H} \right].
    \end{align*}
    In practice, $\bar{H}$ is usually set to $-|\mathcal{A}|$.

  \subsection{Action repeats}
    We call the steps seen from the environment's point of view \emph{environment steps}, and the steps seen from the agent's point of view \emph{agent steps}, respectively.
    Let $T$ be the length of each action repeat. Then, one agent step corresponds to $T$ environment steps, and the agent selects actions $a_0, a_T, a_{2T}, \ldots$ at states $s_0, s_T, s_{2T}, \ldots$.
    As for the actions at the other environment steps, the previously selected actions are repeated, i.e., $a_0=a_1=\ldots=a_{T-1}, \;a_{T}=a_{T+1}=\ldots=a_{2T-1}, \; \ldots$.
    In conventional RL settings, only $s_0, s_{T}, s_{2T}, \ldots$ are used for training, and the data in between are not used.

    There are several merits in repeating the same action in this manner.
    First, it increases the action-gap \citep{NIPS2011_4485} and leads to more stable learning \citep{bellemareIncreasingActionGap2016}.
    This is because clear differences between actions facilitate the correct ranking of actions when the value estimation is uncertain.
    Also, long action repeats can facilitate exploration in some cases \citep{hesselInductiveBiasesDeep2019}.
    For example, when the agent is exploring a maze, it may reach a state that has not been visited before by repeating the same action over and over again and move in one direction for a long time.
    Furthermore, decreasing the frequency of action selection reduces the computational cost \citep{hesselInductiveBiasesDeep2019}.
    This can be a particularly important issue in practical applications of RL.
    Since deep RL spends a sizable amount of time performing inference with deep neural networks,  the action repeat parameter sometimes has to be large to ensure real-time operation.
    In addition, when an agent's suggested action is used by human operators, they cannot switch between actions as quickly as a computer.
    In such a case, the agent has to wait for a long time before choosing the next action.

    The length of each action repeat is usually considered to be a hyper-parameter, and its value can have a significant impact on the performance.
    Braylan et al. \citep{braylanFrameSkipPowerful2015} showed that it is very important to set the appropriate value of action repeats for each task in the Atari 2600 benchmark.
    For example, they conducted experiments with several values of action repeats in Seaquest and achieved the highest score with a very high value of $180$.

    The DeepMind control suite also uses different values of action repeat that lead to high performance for different tasks.
    Therefore, the Dreamer Benchmark \citep{hafnerDreamControlLearning2020}, which fixes the value of action repeats, is considered more difficult than the Planet Benchmark \citep{hafnerLearningLatentDynamics2019}, which chooses an action repeat parameter for each task \citep{kostrikovImageAugmentationAll2020}.

    It is hard to find the optimal value of action repeats for each task, and what is more, the appropriate value can change over time even in one task.
    Thus, Sharma et al. \citep{sharmaLearningRepeatFine2017a} proposed a method called FiGAR, in which an agent learns time abstraction by choosing not only actions but also the number of times to repeat them.
    This method has improved performance on some tasks of Atari 2600 and MuJoCo \citep{todorovMuJoCoPhysicsEngine2012}.


%% file: proposal.tex
\section{Proposed Approach}
  \label{sec:proposal}

  As described in Section \ref{sec:background}, only states $s_0, s_{T}, s_{2T}, \ldots$ are used for the standard policy training, and the data in between are not used.
  We aim to utilize such data to improve the learning process.

  \subsection{Overview}
    Many off-policy reinforcement learning algorithms, including Deep Q-learning and SAC, involve the fitting of Q-functions.
    The proposed method utilizes data during action repeats to train the Q-networks.

\begin{figure}
      \subfloat[transition data between action-decision points]{%
        \includegraphics[width=0.49\textwidth]{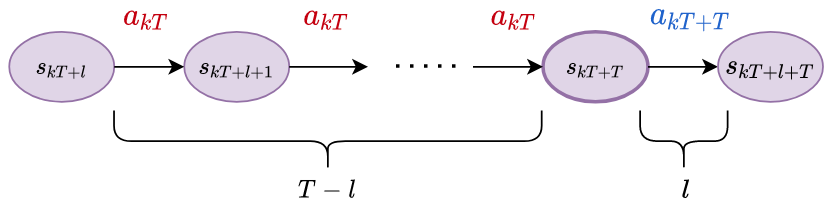}
        \label{fig:repeat-1}
      } \quad
      \subfloat[pseudo-transition data used in our approach]{%
        \includegraphics[width=0.49\textwidth]{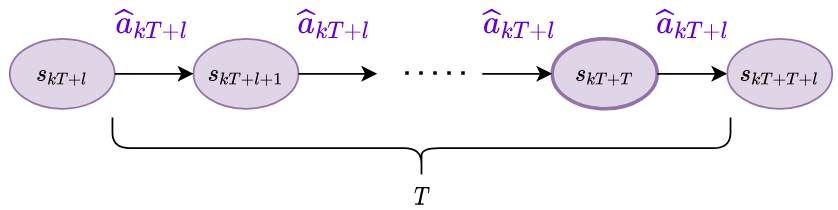}
        \label{fig:repeat-2}
      }
      \caption{Overview of the proposed method. In this case, $l=1$.}
    \end{figure}

    A transition over one agent step between action-decision points can be represented as $(s_{kT+l},a_{kT+l},s_{kT+l+T})$, where $k,l\in\mathbb{N}, 0<l<T$.
    The reason why this transition is not readily available for the training is that, in general, action $a_{kT+l} (= a_{kT})$ is not repeated $T$ times from state $s_{kT+l}$ to $s_{kT+l+T}$.
    As shown in Figure \ref{fig:repeat-1}, the agent selects a new action $a_{kT+T}$ at timestep $t={kT+T}$, so $a_{kT+l}$ is repeated only $T-l \, (<T)$ times from $s_{kT+l}$ unless $a_{kT} = a_{kT+T}$.

    In our approach, we consider that the \textit{pseudo-action} $\hat{a}_{kT+l}$ is repeated $T$ times from the state $s_{kT+l}$ as shown in Figure \ref{fig:repeat-2}.
    This makes the pseudo-transition $(s_{kT+l}, \hat{a}_{kT+l}, s_{kT+l+T})$ available for the training of Q-networks.

    In the rest of this section, we will explain our method in detail in the cases of continuous and discrete action spaces separately.

  \subsection{Continous action space}
    As a simple way to compute pseudo-actions, we use the average of the actions $a_{kT+l}, a_{kT+l+1}, \ldots, a_{kT+l+T-1}$, i.e.:
    \begin{align}
      \hat{a}_{kT+l} = \frac{1}{T} \sum_{t=0}^{T-1} a_{kT+l+t} .
    \end{align}

    If the dynamics of the environment can be locally linearly approximated, we can treat the canonical data and pseudo-data in the same way by using this average of actions.
    The details can be found in Appendix A.

  \subsection{Discrete action space}
    As for discrete action spaces, we cannot simply take the average.
    Therefore, we propose to change the structure of the Q-network so that it can be treated in the same way as that of continuous action spaces.

\begin{figure}
      \subfloat[Q-network for continuous action spaces]{%
        \includegraphics[width=0.31\textwidth]{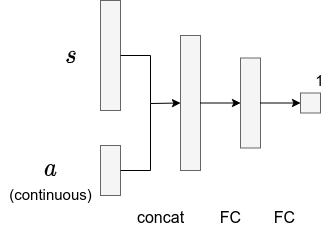}
        \label{fig:models-continuous}
      } \qquad
      \subfloat[Q-network for discrete action spaces]{%
        \includegraphics[width=0.21\textwidth]{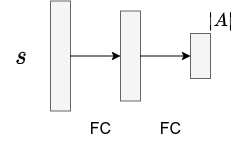}
        \label{fig:models-discrete}
      } \qquad
      \subfloat[Q-network for discrete action spaces used in the proposed method]{%
        \includegraphics[width=0.36\textwidth]{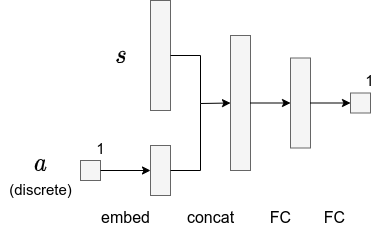}
        \label{fig:models-proposed}
      }
      \caption{Different architectures of Q-networks. "concat", "FC", "embed" mean concatenation, fully connected layer, and embedding layer, respectively.}
    \end{figure}

    In general, the Q-functions of continuous and discrete action spaces are implemented as neural networks shown in Figure \ref{fig:models-continuous} and \ref{fig:models-discrete}, respectively.
    Note that, although this work deals with image observations, we omit convolutional layers from the figures for simplicity.
    As shown in the figures, in continuous action spaces, a single Q-value is output per each input, while in discrete action spaces, Q-values for all actions are output.

    In order to make the network for discrete action spaces similar to that for continuous action spaces, the structure is changed as shown in Figure \ref{fig:models-proposed}.
    An action is converted into an embedding, and the output is a single Q-value.
    The action embeddings are trained jointly with the parameters of the Q-network.
    This allows us to use the average of the embeddings as pseudo-actions.
    Let $e_\theta(a)$ be the embedded representation of action $a$, then the embedding of the pseudo-action $e_\theta(\hat{a}_{kT+l})$ is calculated as
    \begin{align*}
      e_\theta(\hat{a}_{kT+l}) = \frac{1}{T} \sum_{t=0}^{T-1} e_\theta(a_{kT+l+t}) .
    \end{align*}

  \subsection{Implementation}
    The only change in the proposed method is to use the pseudo-data in addition to the canonical data, and it can be easily implemented.
    Algorithm \ref{alg:proposed} shows the pseudo-code to create a mini-batch for the Q-network training on continous control tasks.
    For discrete control tasks, we obtain the action embeddings $e_\theta(a)$ within the Q-network training and take the average of them.

    We treat the canonical data and pseudo-data equally.
    That is, the index $i$ of the transition data in Algorithm \ref{alg:proposed} is sampled uniformly.
    The ratio of the amount of the canonical data to the pseudo-data is then $1$ : $T-1$.
    On the other hand, if only the data that satisfy $i=kT$ are used, the training reduces to the standard training.
    In both cases, the mini-batch size $N$ is set to the same value.

    Since the pseudo-data are obtained based on approximations, it can have a negative impact on the learning process.
    Therefore, it may be beneficial to give more weight to the canonical data.
    We leave this for future work.

    \begin{algorithm}
      \caption{Create a mini-batch for the training of Q-networks}
      \label{alg:proposed}
      \begin{algorithmic}
        \REQUIRE replay buffer $\mathcal{D}$, mini-batch size $N$, action repeat parameter $T$
        \ENSURE mini-batch $\mathcal{B}$
        \STATE Initialize $\mathcal{B} = \{\}$
        \FOR {$k = 1$ to $N$}
          \STATE Sample $(s_i, a_i, r_i, \ldots, s_{i+T-1}, a_{i+T-1}, r_{i+T-1}, s_{i+T}) \sim \mathcal{D}$ \\
            \hspace*{15mm} where $i=kT+l$ \quad ($k\in \mathbb{N}, l \in [0, T)$)
          \STATE Current state $s = s_i$
          \STATE Next state $s^\prime = s_{i+T}$
          \STATE Reward $r = \sum_{t=0}^{T-1} r_{i+t}$
          \STATE Pseudo-action $\hat{a} = \frac{1}{T} \sum_{t=0}^{T-1} a_{i+t}$
          \STATE Add $(s, \hat{a}, r, s^\prime)$ to $\mathcal{B}$
        \ENDFOR
        \RETURN $\mathcal{B}$
      \end{algorithmic}
    \end{algorithm}

%% file: experiment.tex
\section{Experiments}
  \label{sec:experiment}
    We compare the proposed method with the standard training in several environments of OpenAI Gym.
    In all environments, the observations are images and the last four frames are stacked.
    The number of environment steps is fixed to $400$k in all cases.
    All experiments were conducted over five random seeds.
    The details of the neural network models and the hyper-parameters can be found in Appendix B, C.

  \subsection{Continous control tasks}
    We evaluate the proposed method on two continuous control tasks, namely Pendulum and CarRacing.
    In this experiment, we adopt SAC as a reinforcement learning algorithm.
    Although SAC trains the policy network as well as the Q-network, for fair comparison, we do not use the data between action-decision points to train the policy network.

\begin{figure}
      \begin{center}
        \subfloat[Pendulum]{%
          \includegraphics[width=1.0\textwidth]{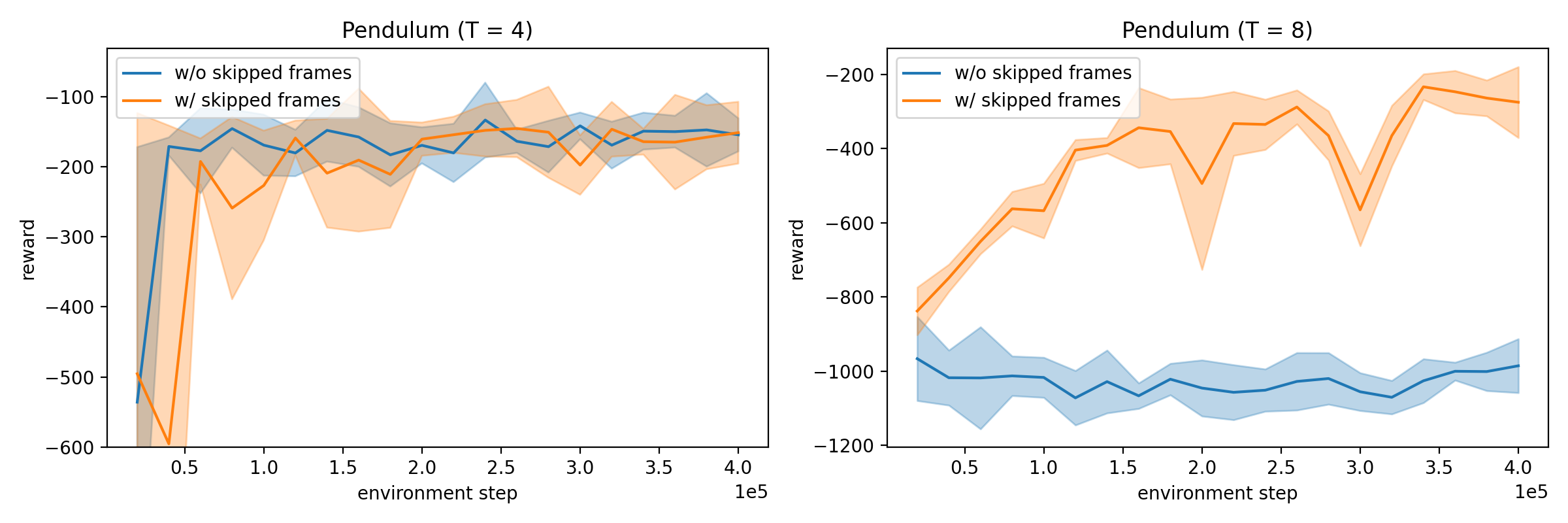}
          \label{fig:pendulum}
        }
      \end{center}
      \begin{center}
        \subfloat[CarRacing]{%
          \includegraphics[width=1.0\textwidth]{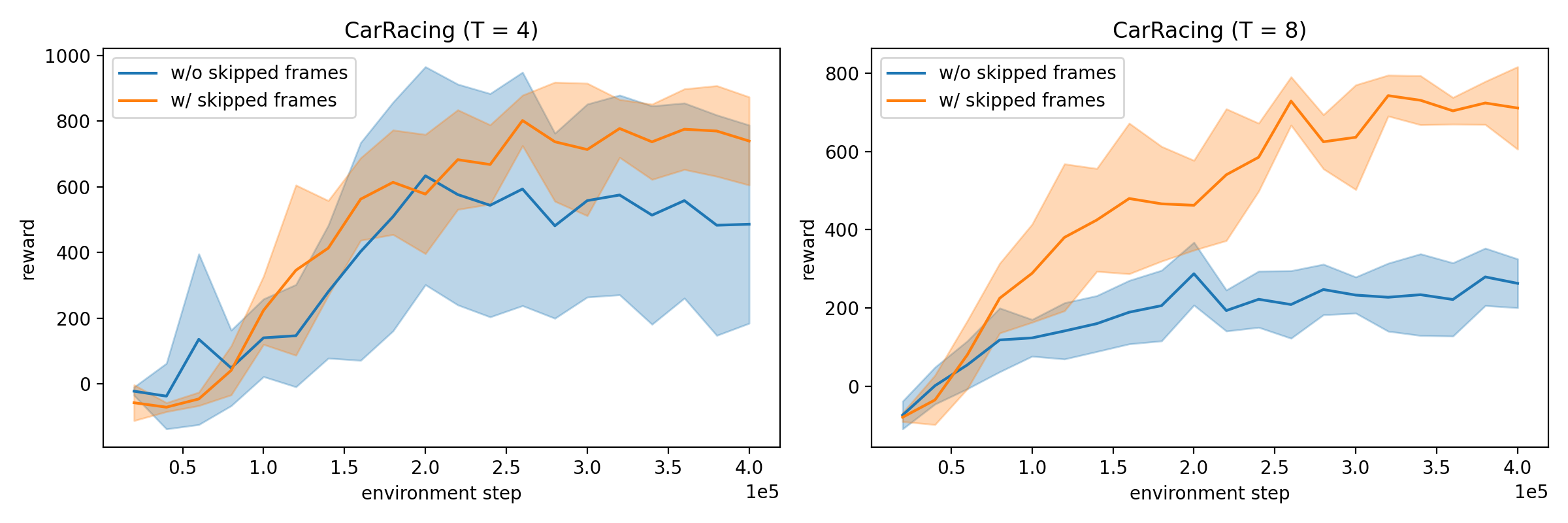}
          \label{fig:racing}
        }
      \end{center}
      \caption{Results on continuous control tasks. Solid lines represent the mean and colored areas represent the standard deviation over five runs.}
      \label{fig:continuous}
    \end{figure}

    The results are shown in Figure \ref{fig:continuous}.
    Compared with the baseline which does not use the data during action repeats, the performance of our method is higher especially in the case of $T=8$.
    This can be attributed to the fact that the baseline can only use $1/8$ of the whole data when $T=8$, while our method can use all of them.
    Furthermore, considering the fact that the standard training completely fails in Pendulum when $T=8$, it is possible that one out of $8$ frames does not have enough information to capture the dynamics of the environment.
    On the other hand, when $T=4$, there is no clear difference in performance for Pendulum, and our method performs slightly better than the baseline for CarRacing.
    In this case, the amount of data discarded in the baseline training is smaller, which leads to the successful policy learning.

  \subsection{Discrete control tasks}
    We evaluate the proposed method on two discrete control tasks, namely Breakout and Freeway.
    In this experiment, we adopt Deep Q-learning as a reinforcement learning algorithm.
    Note that the structure of the Q-network is different from the usual one, and the action embeddings are used in the network, as described in Section \ref{sec:proposal}.
    As a method for exploration, we use the Noisy Network \citep{fortunatoNoisyNetworksExploration2018}.

\begin{figure}
      \begin{center}
        \subfloat[Breakout]{%
          \includegraphics[width=1.0\textwidth]{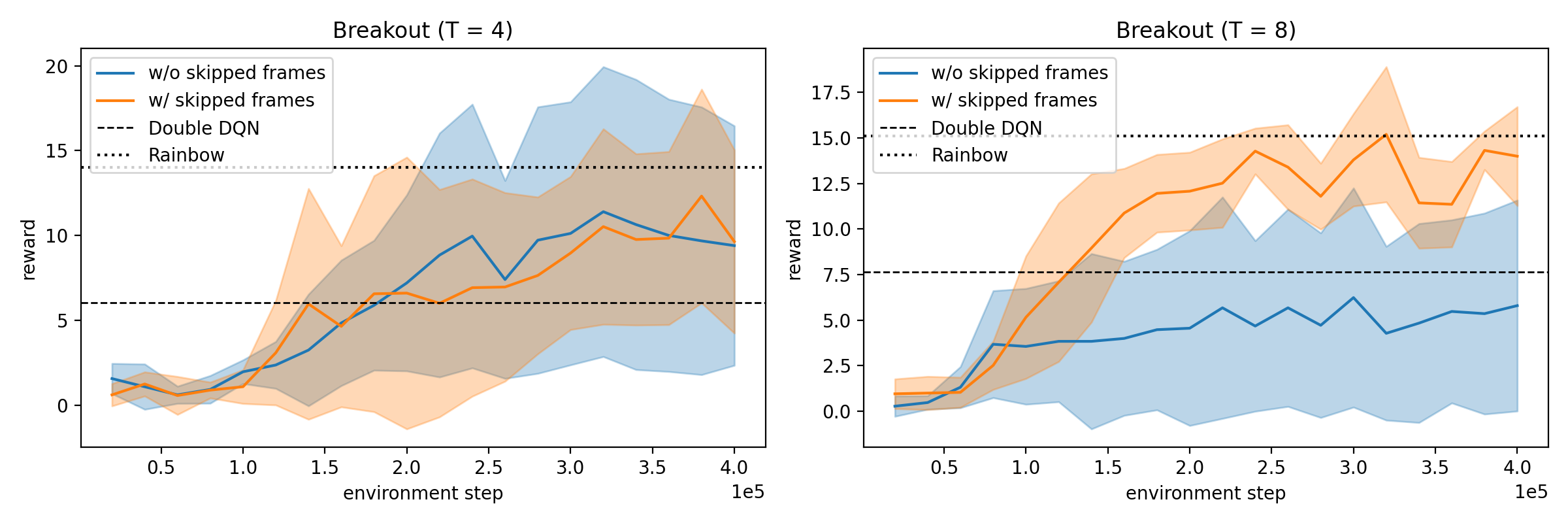}
          \label{fig:break}
        }
      \end{center}
      \begin{center}
        \subfloat[Freeway]{%
          \includegraphics[width=1.0\textwidth]{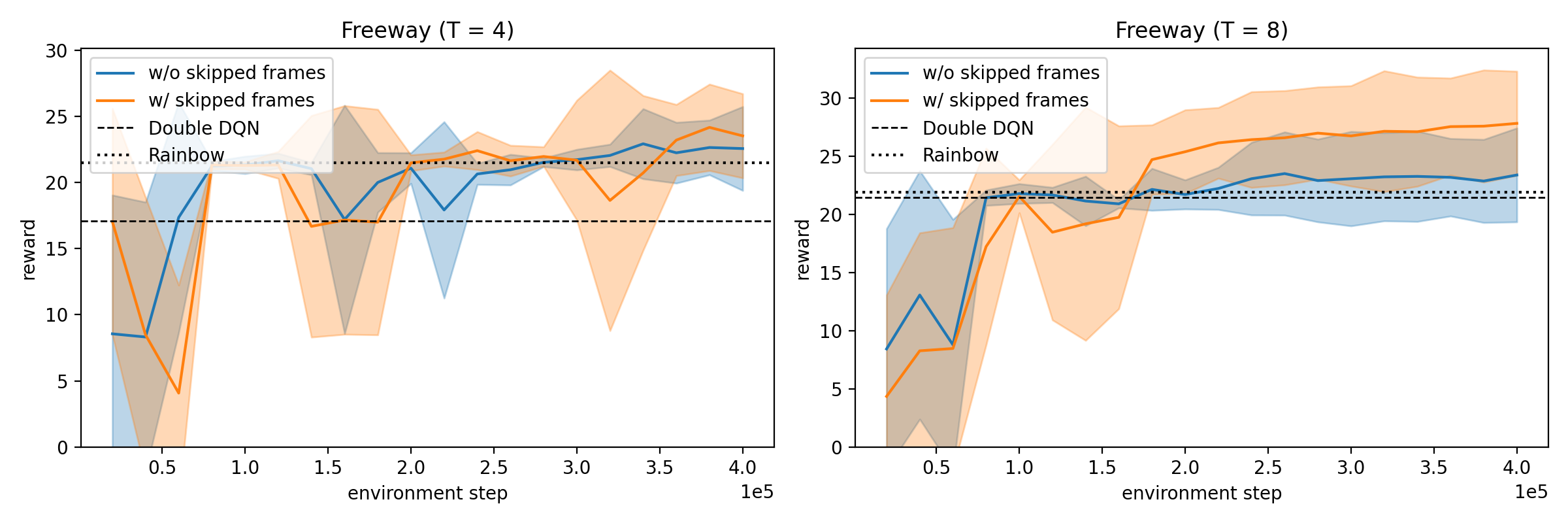}
          \label{fig:freeway}
        }
      \end{center}
      \caption{Results on discrete control tasks. Solid lines represent the mean and colored areas represent the standard deviation over five runs. Dotted lines represent the mean of the final performance of Double DQN and Rainbow.}
      \label{fig:discrete}
    \end{figure}

    The results are shown in Figure \ref{fig:discrete}.
    To make sure that the scores are not significantly reduced by changing the configuration of the Q-networks, we also include the results of Double DQN \cite{vanhasseltDeepReinforcementLearning2016} and Rainbow \cite{hesselRainbowCombiningImprovements2018} with the standard network configuration, trained using the same number of environment steps.
    As in the continuous control tasks, there is no significant difference between our method and the baseline for $T=4$, and our method performs better for $T=8$.

  \subsection{Effects of action repeats}
\begin{figure}
      \centering
      \subfloat[CarRacing]{%
        \includegraphics[width=0.5\textwidth]{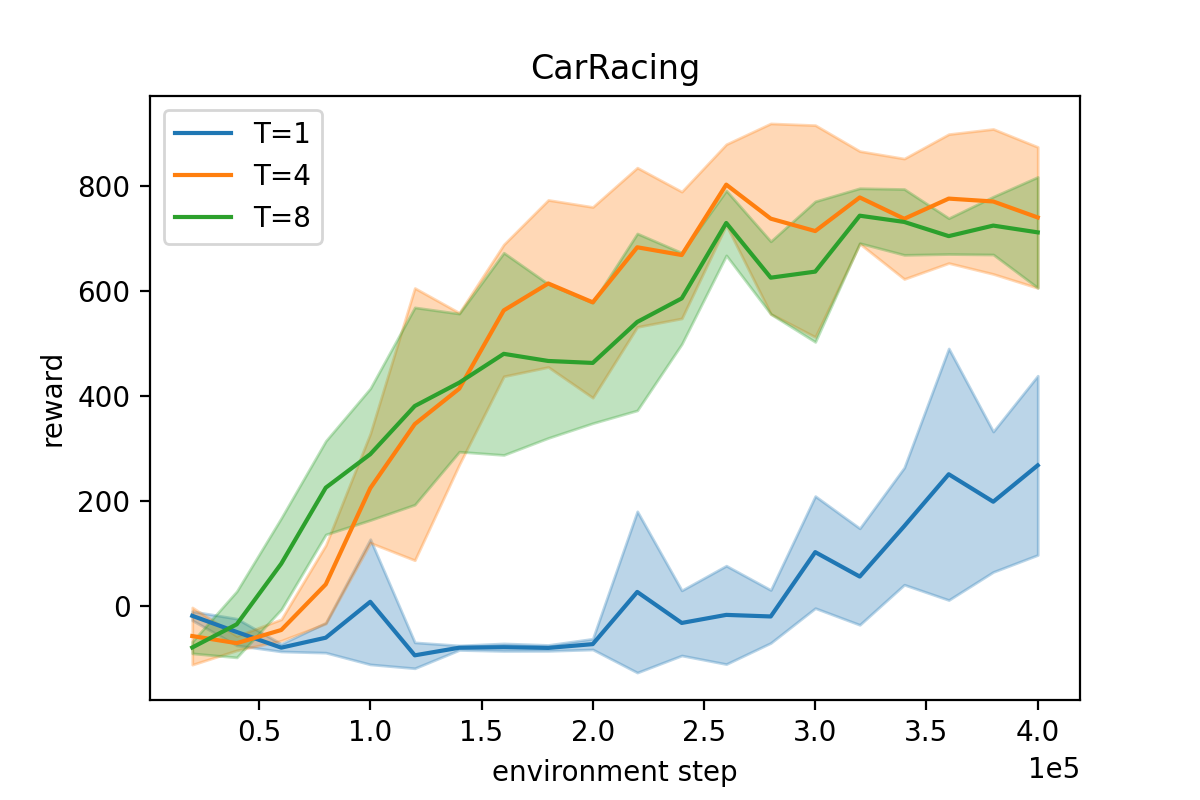}
      }
      \subfloat[Pendulum]{%
        \includegraphics[width=0.5\textwidth]{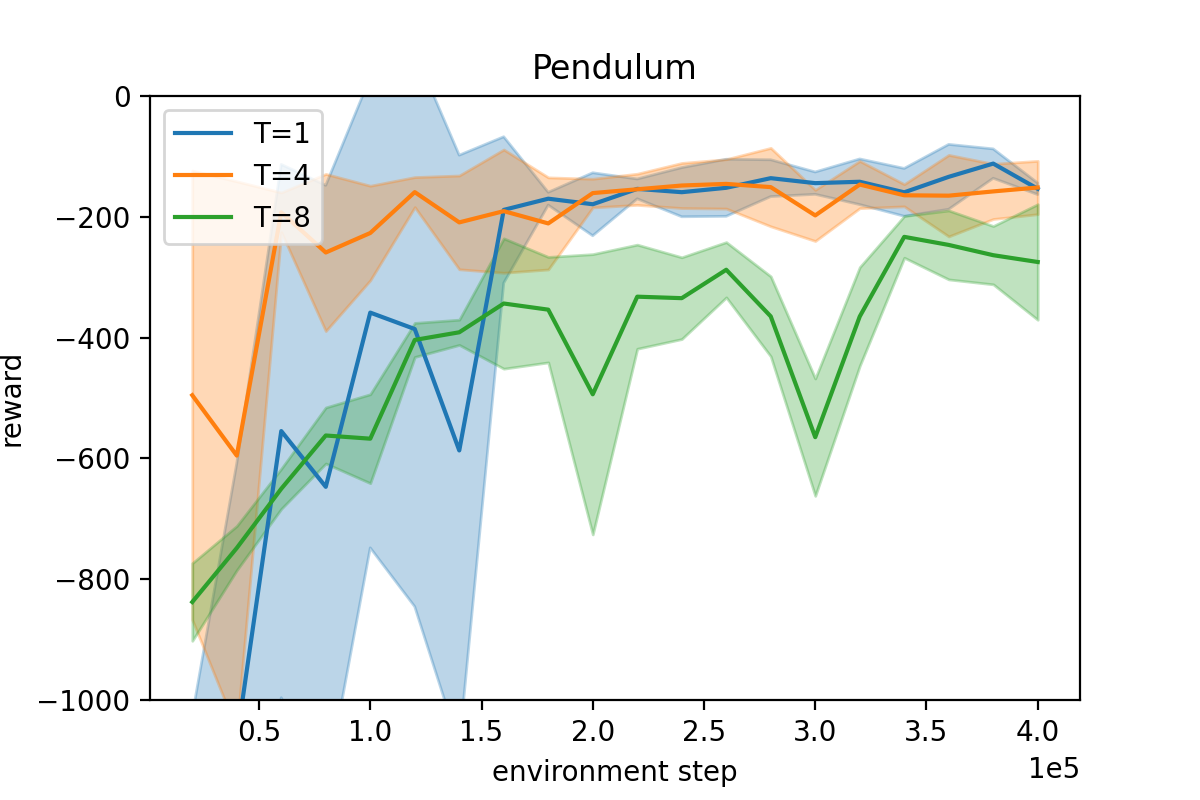}
      } \quad
      \subfloat[Breakout]{%
        \includegraphics[width=0.5\textwidth]{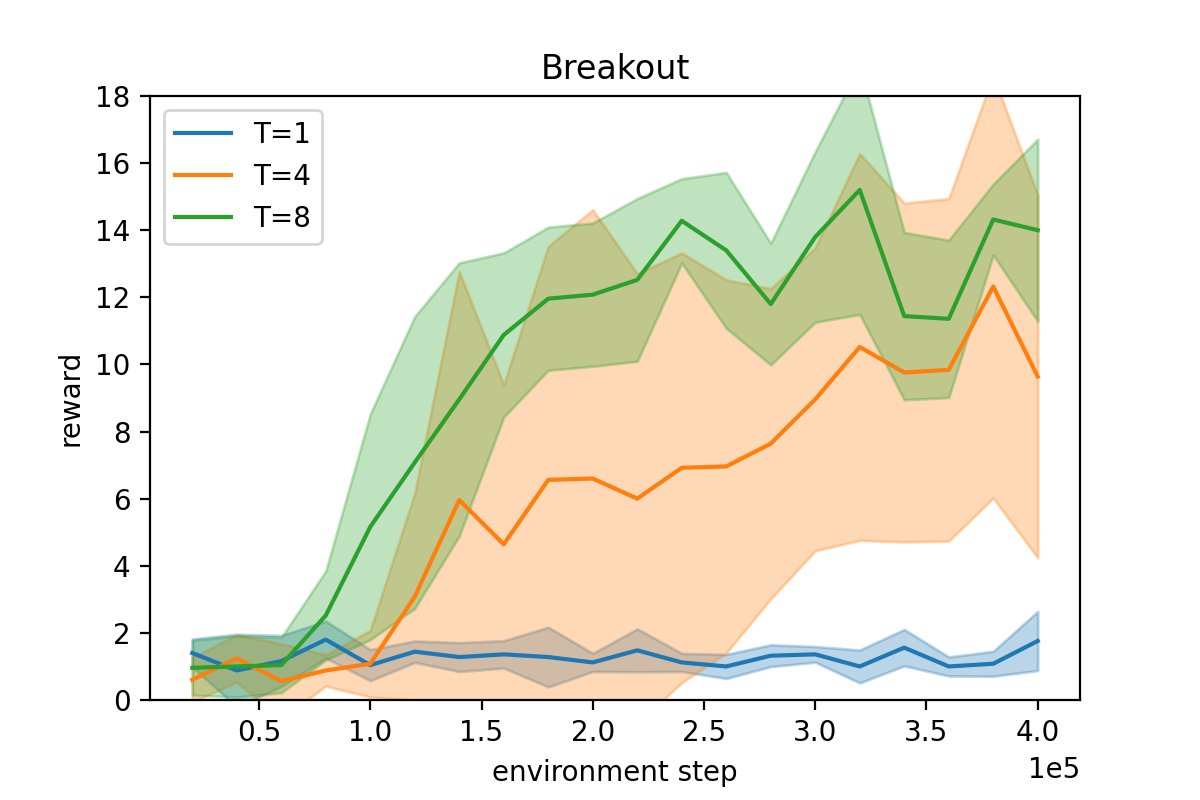}
      }
      \subfloat[Freeway]{%
        \includegraphics[width=0.5\textwidth]{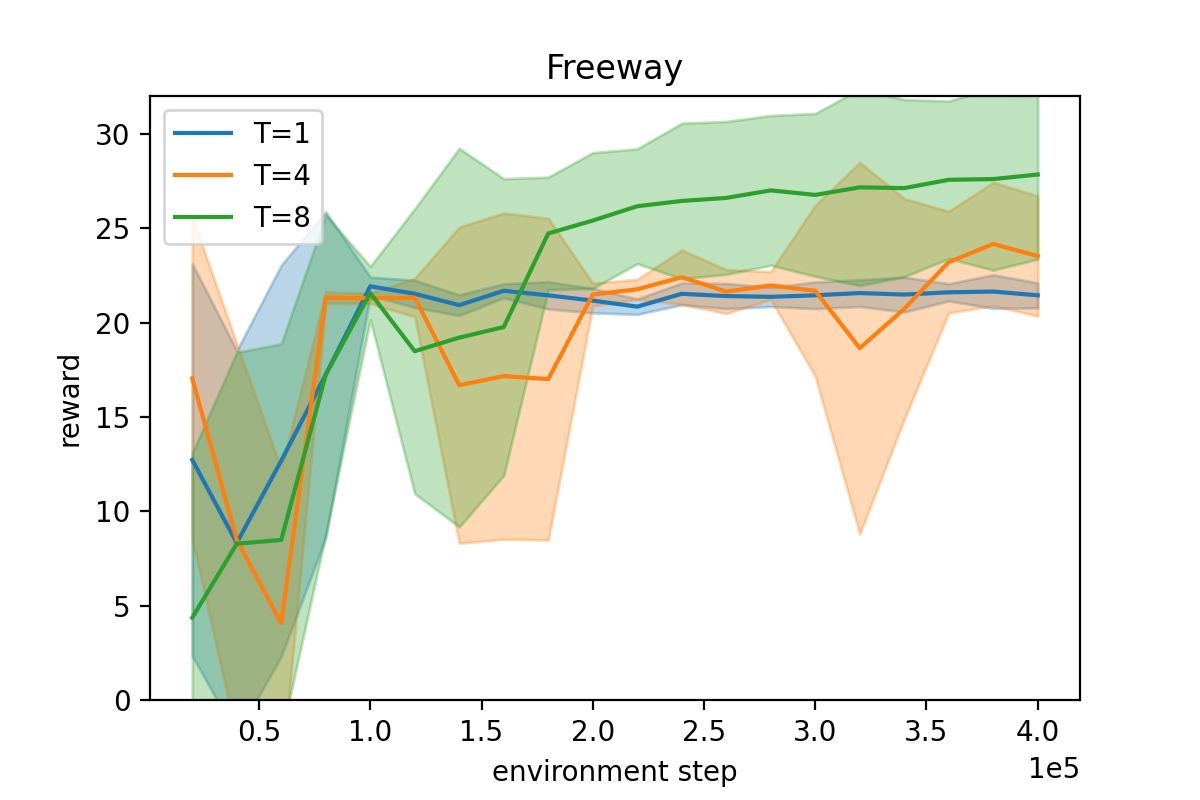}
      }
      \caption{Comparison of different action repeat parameters. Solid lines represent the mean and colored areas represent the standard deviation over five runs.}
      \label{fig:action-repeat-proposed}
    \end{figure}

    We compare the results of our method for $T=1,4,$ and $8$ to evaluate the effects of action repeats.
    Note that for $T=1$, our method is the same as the standard training.
    Results are shown in Figure \ref{fig:action-repeat-proposed}.
    Except for Pendulum, action repeats have a positive impact on the performance.
    This can be attribuited to the increase in the action-gap \citep{bellemareIncreasingActionGap2016} and better exploration \citep{hesselInductiveBiasesDeep2019}, as described in Section \ref{sec:introduction}.
    Thus, action repeats are necessary for many tasks and it is important to utilize the skipped frames.
    For the comparison under the standard training, refer to Appendix D.

  \subsection{Discussions}
    In our experiments, for both continuous and discrete control tasks, our method is superior to the baseline when the interval of action repeats is long, and otherwise, their results are similar.
    While this is a natural outcome, the amount of data available for our method at $T=4$ is four times larger than the baseline, so we may expect that our method would perform better than the baseline even in this case.

    The reason why such results are not clearly shown in the experiments may be that the modeling of pseudo-actions is suboptimal, and the additional data can hurt the learning process.
    Although we simply used the average of the actions in the continuous control tasks, a better representation may be obtained by using an inverse model that infers actions from state changes \citep{pathakCuriosityDrivenExplorationSelfSupervised2017,chandakLearningActionRepresentations2019}.

    Furthermore, for the discrete control tasks, the averages of action embeddings are used as the pseudo-actions, and these representations are obtained purely from the training of the Q-network.
    Therefore, the action representations for these tasks are not directly related to the dynamics of the environment, unlike those for the continuous control tasks.
    If we instead acquire the action embeddings through the training of the transition model, the representations are expected to reflect the dynamics.
    By using these representations, the pseudo-actions can be better modeled and the performance of our method can be improved.


%% file: conclusion.tex
\section{Conclusion}
  \label{sec:conclusion}
  We have proposed a method to utilize the discarded data during action repeats in both continuous and discrete control tasks.
  Our experimental results on the OpenAI Gym environments demonstrate the effectiveness of our method in several configurations.
  We have also confirmed that the longer action repeats lead to higher scores on several tasks, which corroborates some theoretical advantages of action repetition.

  One possible direction of future work is to refine our approach to obtain pseudo-actions in a more sophisticated way.
  As described in Section \ref{sec:experiment}, there is room for improvement in the proposed method and it would further enhance the performance.

  In this paper, we focused on online RL.
  Recently, offline RL, where agents learn a policy without interacting with the environment, has been actively studied \citep{levineOfflineReinforcementLearning2020,agarwalOptimisticPerspectiveOffline2020}.
  Since offline RL assumes static datasets and collecting more transitions is not allowed, augmenting the data like the proposed method can be important.
  Our method is readily applicable to offline RL as long as the transitions between action-decision points are stored during data collection, and extending our approach to such settings should also be an interesting direction for future research.


%% file: broader_impact.tex
\section*{Broader Impact}
  \label{sec:broader-impact}

  Since action repetition is commonly used in deep RL, our proposed method is applicable to a wide range of tasks.
  Furthermore, our method can be combined with many RL algorithms on continuous and discrete action spaces.

  There are some particular cases where our method would have practical benefits.
  Firstly, it can boost the sample efficiency on tasks where the data collection is costly.
  Also, as shown in our experiments, some tasks are hard to solve with long action repeats, and we need to utilize the intermediate states in these cases.
  In addition, high action frequency is sometimes infeasible due to communication delays and inference time.
  We assume that robotic control is a good example of tasks with these properties.
  The potential risk of our method is that it may impair the performance especially when the linear approximation of the dynamics does not hold.
  To mitigate it, we have to improve our methodology and to investigate the conditions under which our approach works well.

  In spite of its popularity, there has not been much research on the action repeat itself.
  Further research on this simple yet powerful technique is expected, and more realistic settings of action repeats should be adopted in RL experiments with practical applications in mind.
  Also, there are some other techniques used in RL, such as stacked frames and normalized advantages, so we encourage the research to understand the their impacts and to improve them.

%% file: appendix.tex
\appendix

\section{Details about pseudo-actions}
  \label{app:about-pseudo-actions}
  As described in the Section 3, as for the transitions between action-decision points, repeating action $a_{kT+l} (= a_{kT})$ $T$ times from state $s_{kT+l}$ does not lead to $s_{kT+l+T}$.
  Therefore, we cannot treat these transitions as regular training data.
  In contrast, as shown below, repeating pseudo-action $\hat{a}_{kT+l}$ $T$ times from state $s_{kT+l}$ leads to the state which is approximately equal to state $s_{kT+l+T}$.

  Here, we consider a continuous time setting.
  Suppose that the agent's state $x(t)$ and action $u(t)$ are in a relationship as below
  \begin{align*}
    x(t+T) = \int_{t}^{t+T} f(x(t^\prime), u(t^\prime)) \, dt^\prime
  \end{align*}
  where $f$ is an unknown dynamics.
  Assuming that the state does not change drastically in $t\leq t^\prime \leq t+T$,
  \begin{align*}
    x(t+T) \approx \int_{t}^{t+T} f(x(t), u(t^\prime)) \, dt^\prime.
  \end{align*}
  Here, we assume that
  \begin{align*}
    u(t) = \begin{cases}
      u_1 & (t \leq t+pT) \\
      u_2 & (t > t+pT)
    \end{cases}
  \end{align*}
  where $p\in\mathbb{R}, 0<p<1$.
  In other words, the agent repeats an action $u_1$ until $t+pT$, and repeats a new action $u_2$ afterwards.
  Then, 
  \begin{align*}
    x(t+T) &\approx \int_{t}^{t+pT} f(x(t), u_1) \, dt^\prime + \int_{t+pT}^{t+T} f(x(t), u_2) \, dt^\prime \\
    & = pT \, f(x(t), u_1) + (1-p)T\, f(x(t), u_2) .
  \end{align*}

  Let $\hat{u}$ be the average of actions
  \begin{align*}
    \hat{u} = p \, u_1 + (1-p)\, u_2 .
  \end{align*}
  Here, $\hat{u}$ corresponds to the pseudo-action of our method.
  Note that $u_1$ and $u_2$ can be expressed as
  \begin{align*}
    u_1 = \hat{u} + (1-p) \, (u_1 - u_2), \quad u_2 = \hat{u} - p \, (u_1 - u_2) .
  \end{align*}
  Then, a first-order approximation with respect to the action yields
  \begin{align*}
    x(t+T) &\approx pT \, f(x(t), \hat{u} + (1-p)\,(u_1-u_2)) + (1-p)T \, f(x(t), \hat{u} - p\,(u_1-u_2)) \\
    &\approx pT \, \left\{ f(x(t), \hat{u}) + (1-p)\,(u_1-u_2)\,\frac{\partial f}{\partial u}(x(t), \hat{u}) \right\} \\
    &\hspace{3mm} + (1-p)T \, \left\{ f(x(t), \hat{u}) - p\,(u_1-u_2)\,\frac{\partial f}{\partial u}(x(t), \hat{u}) \right\} \\
    &= T f(x(t), \hat{u}) \\
    &= \int_{t}^{t+T} f(x(t), \hat{u}) \, dt^\prime \approx \int_{t}^{t+T} f(x(t^\prime), \hat{u}) \, dt^\prime .
  \end{align*}
  As a result, replacing $u(t)$ with the constant $\hat{u}$ roughly does not change the next state $x(t+T)$.

\section{Neural Network Models}
  \label{app:neural-network-models}
  As described in Section 3, our method makes the network for discrete control tasks similar to that of continuous control tasks.
  For this purpose, we adopt the same network structure for the image encoding network and the Q-network.

  \subsection{Encoder Network}
    The structure of the encoder network is based on SAC+AE \citep{yaratsImprovingSampleEfficiency2020}.
    This model has been adopted by several studies that use SAC to learn policy from image observations \citep{srinivasCURLContrastiveUnsupervised2020,kostrikovImageAugmentationAll2020}.
    The encoder consists of four convolutional layers with $3 \times 3$ kernels and $32$ channels.
    We use ReLU as an activation function.
    The stride of the first layer is set to $2$, and the subsequent layers have the strides of $1$.
    The output of the convolutional layers is normalized by LayerNorm \citep{baLayerNormalization2016} and is converted into a $50$ dimensional vector by a fully connected layer, followed by a $\tanh$ activation.

  \subsection{Q-network and policy network}
    Both the Q-network and the policy network consist of three fully connected layers, and the hidden dimension is set to $256$.
    We use ReLU as an activation function.
    The Q-network takes an output of the encoder network and an action (or an action embedding) and outputs a single Q-value.
    The dimension of embeddings for discrete action spaces is set to $8$.
    The policy network, which is used only for the continuous control tasks, takes an output of the encoder network and outputs the mean and variance of a diagonal Gaussian distribution representing the policy.

\section{Other Hyper-parameters}
  \label{app:other-hyper-parameters}
  Table \ref{table:hyper-parameters} shows the hyper-parameter settings.
  For discrete control tasks, we use Double DQN \citep{vanhasseltDeepReinforcementLearning2016} to stabilize the training.
  To deal with different reward scales over tasks, we adopt PopArt \citep{hesselMultitaskDeepReinforcement2018} for continuous control tasks and clipped rewards ($[-1, 1]$) for discrete control tasks.

  \begin{table}[H]
    \caption{Hyper-parameters used for experiments}
    \label{table:hyper-parameters}
    \begin{center}
      \begin{tabular}{lcc}
        \hline
        & Continous control tasks & Discrete control tasks \\
        \hline \hline
        Algorithm & SAC & Double DQN \\
        Environment steps & 400k    & 400k \\
        Image size & 84 $\times$ 84 & 84 $\times$ 84 \\
        Frame stack & 4 & 4 \\
        Environment steps per parameter update  & 4 & 4 \\
        Optimizer & Adam    & Adam \\
        Learning rate & 0.001   & 0.0003 \\
        Discount factor per environment step $\gamma$ & $0.99^{1/4}$ & $0.99^{1/4}$ \\
        Target update rate $\tau$ & 0.005 & 0.005 \\
        Actor update frequency & 2 & - \\
        Mini-batch size $N$ & 32 & 64 \\
        Replay buffer size & Unbounded & Unbounded \\
        Minimum replay size for training & 500 & 500 \\
        Maximum environment steps per episode & Unbounded & $108$k \\
        \hline
      \end{tabular}
    \end{center}
  \end{table}

\section{Effects of action repeats under the standard training}
  \label{app:action-repeat-standard}

\begin{figure}[H]
      \centering
      \subfloat[CarRacing]{%
        \includegraphics[width=0.5\textwidth]{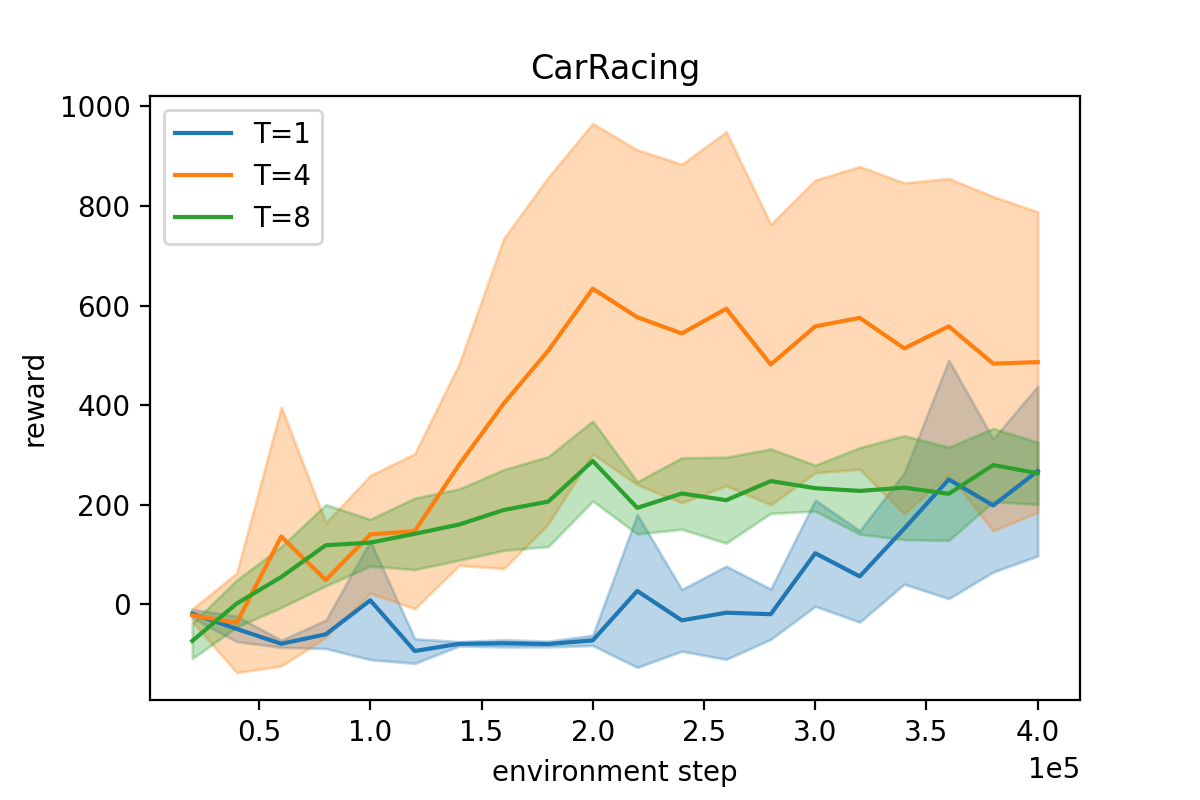}
      }
      \subfloat[Pendulum]{%
        \includegraphics[width=0.5\textwidth]{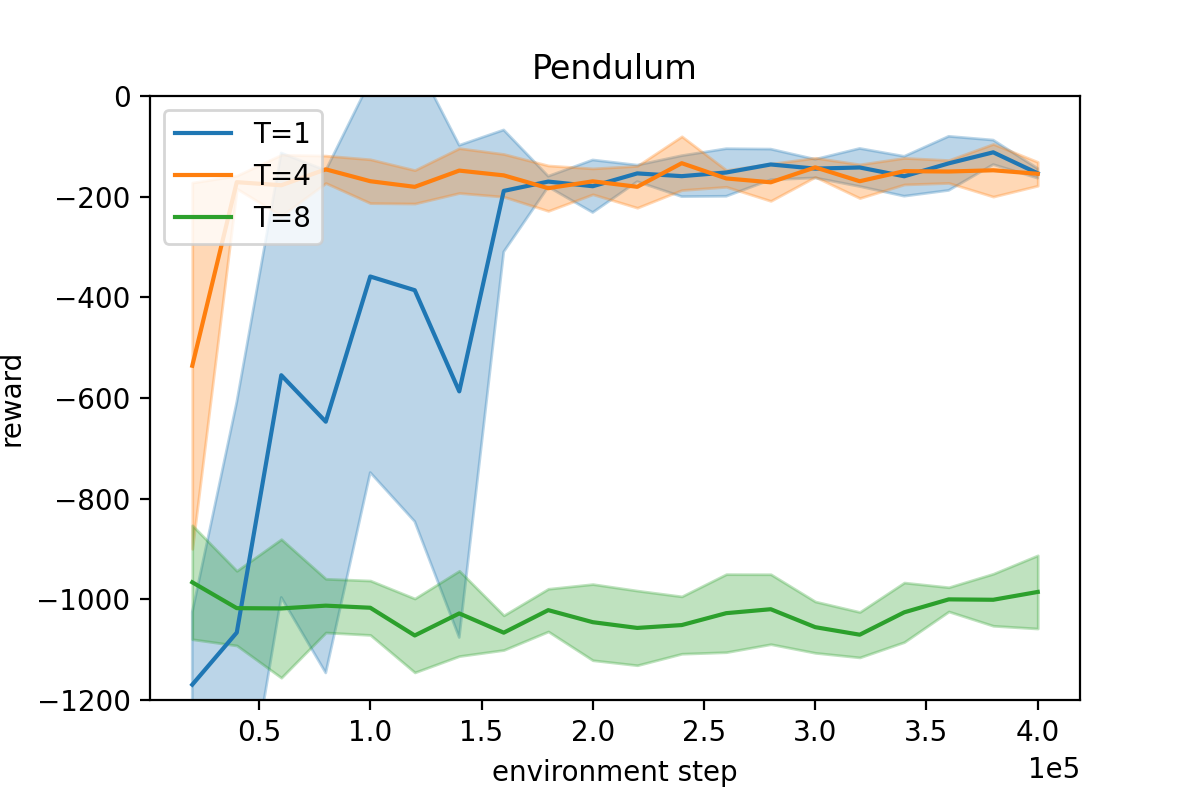}
      } \quad
      \subfloat[Breakout]{%
        \includegraphics[width=0.5\textwidth]{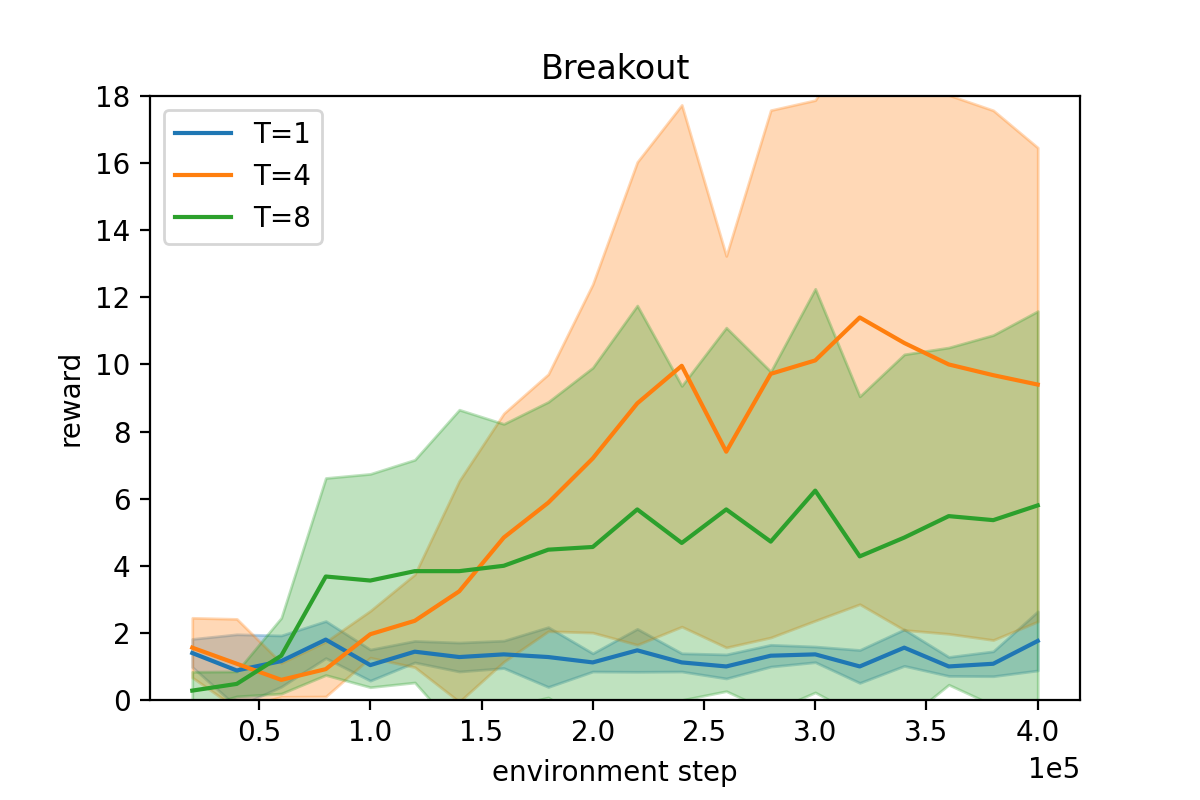}
      }
      \subfloat[Freeway]{%
        \includegraphics[width=0.5\textwidth]{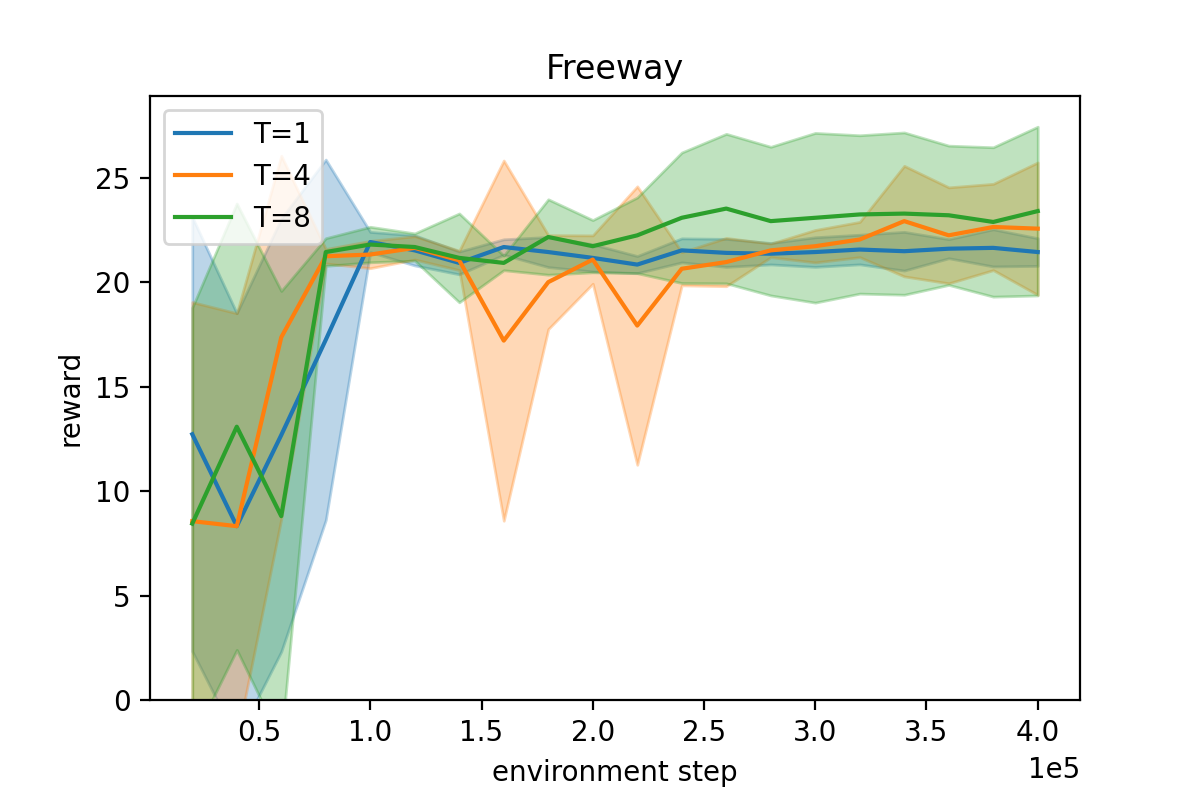}
      }
      \caption{Comparison of different action repeat parameters under the standard training. Solid lines represent the mean and colored areas represent the standard deviation over five runs.}
      \label{fig:action-repeat-standard}
    \end{figure}

  Figure \ref{fig:action-repeat-standard} shows the results of the baseline with different action repeat parameters.
  While action repeats have a positive effects in some cases, the highest value ($T=8$) generally performs much worse than the medium value ($T=4$) due to the lack of training data.
  This can be mitigated by our method as shown in Figure \ref{fig:action-repeat-proposed}.

  